%% file: acl_latex.tex
\pdfoutput=1
\documentclass[11pt]{article}

\usepackage[]{acl}

\usepackage{times}
\usepackage{latexsym}

\usepackage[T1]{fontenc}
\usepackage[utf8]{inputenc}

\usepackage{microtype}

\usepackage{inconsolata}

\usepackage{graphicx}
\usepackage{booktabs}
\usepackage{amsmath}
\usepackage{multirow}
\usepackage{xspace}
\usepackage{setspace}
\usepackage[flushleft]{threeparttable}

\def\chunk{{\tt chunk}\xspace}
\def\prefix{{\tt prefix}\xspace}
\def\full{{\tt full}\xspace}
\def\OOTB{{\tt OOTB}\xspace}
\def\randpos{{\tt RandomPos}\xspace}

\title{Extending Input Contexts of Language Models through Training on Segmented Sequences}

\author{Petros Karypis \\ UC San Diego \\  pkarypis@ucsd.edu
        \AND
        Julian McAuley \\ UC San Diego  \\  jmcauley@ucsd.edu \And
        George Karypis \\ University of Minnesota \\  karypis@umn.edu}

\begin{document}
\maketitle

\input{sections/1_abstract}
\input{sections/2_introduction}

\input{sections/3_relatedworks}

\input{sections/4_method}
\input{sections/5_experiment}

\input{sections/6_results}
\input{sections/7_discussion}

\input{sections/8_conclusion}
\input{sections/9_limitations}
\input{sections/10_ethics}

\bibliography{anthology,custom}

\appendix
\include{sections/11_appendix}

\end{document}

%% file: sections/1_abstract.tex
\begin{abstract}
Effectively training language models on long inputs poses many technical challenges. 
%%%%%% OLD  %%%%%%
% Traditionally, models are limited to the input size on which they were trained. More recently, relative positional embeddings have been used due to their ability to extrapolate to longer sequences than they were trained on but their practical extrapolation ability has been limited. 
%%%%%% NEW  %%%%%%
As a cost consideration, languages models are pretrained on a fixed sequence length before being adapted to longer sequences. 
%%%%%%%%%%%%
We explore various methods for adapting models to longer inputs by training on segmented sequences and an interpolation-based method for extending absolute positional embeddings. We develop a training procedure to extend the input context size of pretrained models with no architectural changes and no additional memory costs than training on the original input lengths. By sub-sampling segments from long inputs while maintaining their original position the model is able to learn new positional interactions. Our method benefits both models trained with absolute positional embeddings, by extending their input contexts, as well as popular relative positional embedding methods showing a reduced perplexity on sequences longer than they were trained on. We demonstrate our method can extend input contexts by a factor of $4\times$ while improving perplexity.
\end{abstract}

%% file: sections/2_introduction.tex
\section{Introduction}

% Out of the box Transformers~\cite{Vaswani2017AttentionIA} are not able to model the sequential nature of text. To address this, positional embeddings are often added to the input to capture this information. 
Transformer-based models~\cite{Vaswani2017AttentionIA} capture sequence information through positional embeddings (PE). There are two types of PEs: absolute and relative. Absolute positional embeddings (APE) learn a separate embedding for each position in a sequence; these embeddings are added to the input of the first layer. Relative positional embeddings (RPE) encode the relative distance between positions, often by weighting attention score of positions further away less. 
% Originally~\cite{Vaswani2017AttentionIA} proposed two options; learnable positional embeddings and fixed sinusoidal relative positional embeddings. Relative positional embeddings usually penalize the attention scores as a function of a inputs relative position to others in the sequence. The intuition behind this tokens near by in a sequence have a greater contextual impact then those further away. 

The ability for models to process long sequences efficiently is of growing importance as models become 
 more capable. Increased input context allows for more complex in-context learning examples~\cite{Li2023InContextLW, Sun2023PEARLPL}. 
% Long input contexts
Additionally, they allow for question answering and summarization over scientific papers and patents~\cite{Dasigi2021ADO, Koh2022AnES, Sharma2019BIGPATENTAL}. 
% More recently, repository level code generation tasks~\cite{Liu2023RepoBenchBR}.
Due to RPE's positional information only being a function of relative distance these methods can be applied to any input sequence length. In practice, popular RPE methods fail to generalize to sequences longer than they were trained on. Furthermore, self-attention's memory cost is quadratic meaning training on long sequences becomes prohibitively expensive as the sequence length grows. 

% In this paper we re-evaluate different positional embeddings ability to extrapolate. Borrowing from ~\cite{Press2021TrainST}, we define a extrapolation as a model's ability to maintain its performance on sequences longer than they were trained on. In our re-evaluation we find that interpolation is a viable method for enabling extrapolation of absolute positional embeddings. We propose a simple approach for extending the effective input context by training on segmented sequences sampled from sequences longer than the model's original input context. Results on models trained with three separate positional embedding methods show this approach performs well regardless of the choice of positional embeddings.
In this work, we study the problem of extending the input context of pre-trained decoder-only transformer-based models, considering those that use either absolute or relative positional embeddings.
% that use either absolute or relative positional embeddings. 
We show that an interpolation-based approach allows APE models to extrapolate to sequence lengths longer then they were trained on---matching or outperforming the extrapolation ability of RPE methods like ALiBi~\cite{Press2021TrainST} and RoPE~\cite{Su2021RoFormerET}. To further improve the ability of these models to take advantage of the longer input context, we present resource-efficient methods that continuously pre-train APE- and RPE-based models on carefully sampled segmented sub-sequences of long sequences. Doing so simulates training on long sequences while remaining within a fixed input length. This allows the models to efficiently learn the embeddings of the newly created absolute positions or the relative embeddings associated with the longer pairwise distances.

\begin{figure*}[t]
    \centering
    \includegraphics[width=0.7  \textwidth]{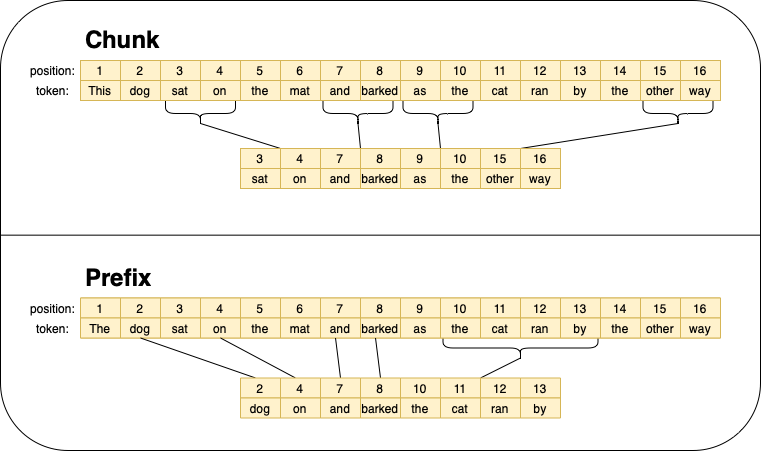}
    \caption{Visualization of our various segment-based methods. We sub-sampling tokens from the original sequence while maintaining the original positions.}
    \label{fig:fcm}
\end{figure*}

We experiment with models trained with APEs, RoPE, and ALiBi to verify our method improves the extrapolation performance independent of the choice of positional embeddings. Results show that interpolating the embedding matrix of absolute positional embeddings without any additional training allows for extrapolation to sequences $5\times $ the original input context. Furthermore, our segment-based methods are able to increase the extrapolation ability of all positional embedding approaches. 
% While ALiBi has been shown to extrapolate to longer sequences our method further improves its ability to do so. 
When applied to APEs this method achieves 87\% the performance of training on sequences twice as long at no extra memory footprint. 

The paper is organized as follows: first, we conduct a review of various existing literature that motivated our approach. Second, we formally define the problem of length extrapolation and propose our methods for efficiently extending a model's input context. Third, we provide a detailed breakdown of our experimental setup and methodology to enable reproducibility. Finally, we present our results along with a thorough discussion and analysis.

%% file: sections/3_relatedworks.tex
%========================================================================
\section{Related Work}
\subsection{Positional embeddings} 
%========================================================================
Language is inherently sequential and Transformers are positional-agnostic, to account for this, positional information is often introduced to the architecture. The original authors~\citet{Vaswani2017AttentionIA} suggested adding a positional embedding to the input of the first layer and offered two methods, absolute positional embeddings and sinusoidal embeddings. Absolute positional embeddings consist of a learnable embeddings matrix where each embedding corresponds to a position. While common, this method has an important limitation: it only allows for a fixed maximum input length determined during training. Sinusoidal embeddings did not have this limitation but performed worse in practice and the relative embeddings that came after were difficult to parallelize~\cite{Shaw2018SelfAttentionWR} leading to APEs being the \textit{de facto} method in early models, eg. BERT~\cite{Devlin2019BERTPO} and GPT-3~\cite{Brown2020LanguageMA}.

To address the limited input context size of APE researchers explored other relative positional embedding methods~\cite{Chi2022KERPLEKR, Wennberg2021TheCF, Likhomanenko2021CAPEER, Haviv2022TransformerLM}. Most notable are rotary embeddings (RoPE)~\cite{Su2021RoFormerET}, T5~\cite{Raffel2019ExploringTL}, and ALiBi~\cite{Press2021TrainST}. RoPE rotates the query and the key embeddings as a function of their position; this method allowed for easier parallelization compared to previous relative embeddings. T5 bias~\cite{Raffel2019ExploringTL} adds a positional embedding for each relative distance instead of absolute position. ALiBi subtracts a linear bias from the query-key matrix product in the  attention calculations. 
% Both methods apply these operations at every layer. 
While T5 bias extrapolated to long contexts well it is too inefficient to scale, taking twice as long to train as sinusoidal~\cite{Press2021TrainST}. RoPE and ALiBi have been widely adopted in various LLMs with LLaMA~\cite{Touvron2023LLaMAOA}, GPT-J~\cite{gpt-j}, and PaLM~\cite{Chowdhery2022PaLMSL} using RoPE and BLOOM~\cite{Scao2022BLOOMA1} using ALiBi.

%========================================================================
\subsection{Length generalization}
%========================================================================

The choice of positional embeddings (PE) has been documented to be one of the leading factors in a Transformer based model's ability to generalize to variable sequence lengths. The authors of ALiBi~\cite{Press2021TrainST} identified that RoPE and sinusoidal embeddings failed to generalize on sequence lengths greater then those they were trained on. Numerous new positional embedding methods with more favorable length generalization abilities have been proposed~\cite{Sun2022ALT, Chi2022KERPLEKR, Li2023FunctionalIF} but these are required to be incorporated during pre-training.  

There is a sizable body of work on methods for extending the input context of language models pre-trained with RoPE~\cite{Chen2023ExtendingCW, Jin2024LLMML, Peng2023YaRNEC, Ding2024LongRoPEEL}. These approaches map the positional information of long sequences into ranges seen during training through positional interpolation. In practice, these methods requires fine-tuning the models on long sequences to adjust to the new granularity of relative positional distance which is computational expensive. 
% More recently,~\cite{Kazemnejad2023TheIO} studied the length generalization ability of various PE methods on a set of auto-regressive tasks. They concluded that Transformers without any positional information generalized the best.

%========================================================================
\subsection{Computationally efficient training} 
%========================================================================

Numerous works have explored efficiency based modifications to the standard Transformer architecture~\cite{Xiong2021NystrmformerAN, Choromanski2020RethinkingAW, Kitaev2020ReformerTE, Qiu2019BlockwiseSF}. These methods  either modify the base architecture or rely on fast self-attention approximations.

While these methods all aim to reduce the memory cost of the Transformer architecture and allow for training on longer sequences, our work is orthogonal to these methods.  Our approach can be used in conjunction with these existing methods since we do not rely on any specific architecture. We instead change the positional information of the input sequences.  

%========================================================================
\subsection{Sparse input sequences}
%========================================================================

A number of works have explored training language models on sparse inputs. 
% It has been demonstrated that encoder language models trained with a masked language modeling objective achieve better performance when using a masking rate of 40\% instead of the 15\% BERT used~\cite{Wettig2022ShouldYM}. Furthermore, their results showed that masking 80\% of the input led to only a 5\% drop in performance during fine-tuning.  
APEs have been shown to overfit to certain positions. To address this, ~\citet{Kiyono2021SHAPESA} proposed randomly padding or offsetting the positions during fine-tuning. This simple method led to better downstream performance on question answering and machine translation~\cite{Tao2023AFE} and general length extension~\cite{zhu2023pose, Ruoss2023RandomizedPE}.
% Of these methods, the one most relevant to our work proposed 
Another work proposed Forgetful Causal Masking (FCM)~\cite{Liu2022FCMFC}, a simple modification to the next token prediction task with a randomly selected fraction of previous tokens masked out. They demonstrated this method led to improvements in both few-shot and fine-tuned performance compared to standard causal masking. Most similar to ours, RandomPos~\cite{Ruoss2023RandomizedPE} proposed sampling randomized, ordered positional embeddings to replace the sequential positional embeddings normally used. They sampled from a range of absolute positions much longer than the input sequence length. Results demonstrated this led to an increase in extrapolation performance. The authors argued this was due to exposure to longer relative pair-wise distances than those normally seen during training. 

These results indicate that not only can language models  be trained with heavily obfuscated sequences but can also benefit from doing so in some cases. This idea is the intuition behind our method.

%% file: sections/4_method.tex
\section{Methods}
\label{sec:methods}
% \begin{figure*}[t]
%     \centering
%     \includegraphics[width=1.0\textwidth]{emnlp2023-latex/images/token_fig.png}
%     \caption{Visualization of our various sparsification methods REMOVE SHAPE.}
%     \label{fig:fcm}
% \end{figure*}

%========================================================================
%\subsection{Notations, definitions, and problem statement}
%\label{ssec:def_notation}
%========================================================================

There are three reasons that motivate this work. First, there exist numerous
high-quality pre-trained models 
% that use either absolute or relative positional encoding schemes 
whose input context is limited to 1K--2K tokens. Extending the input
context of these models will further increase their applicability. Second, even
though methods that rely on relative positional embeddings can operate on input
contexts that are longer than what they were trained on, their out-of-the-box
extrapolation performance is not good~\cite{Press2021TrainST}. Third, due to the quadratic complexity of self-attention and the linear compute/memory complexity of transformers w.r.t.~sequence length, direct training on long input contexts is resource intensive. This limits the input context that we can directly train on.

\subsection{Problem Statement}

Let $p_\theta$ be a transformer-based language model trained to maximize the
next-token-probabilities over a set of sequences $\cal D$ of length $L_t$; i.e.,
\begin{equation}
  \arg\max_\theta \sum_{\mathbf{x}\in \cal D} \sum_i^{L_t} \log p_\theta (x_{i} | x_{<i}).
\end{equation}
We will refer to $L_t$ as the model's \emph{training input context length}.
% and assume
% that $p_\theta$ was trained using either an absolute or a relative positional
% encoding method.
% $P_e$.

We define \emph{extrapolation} as the language model's ability to improve its
next-token-prediction by using input contexts that are longer than those it trained
on. Specifically, for $k>L_t$, we will consider that a model can extrapolate
successfully if
\[ \sum_{i\ge k} \log p_\theta(x_i|x_{>k})
>
\sum_{i\ge k} \log p_\theta(x_i|x_{>L_t}),\]
where $p_\theta(x_i|x_{>j}) = p_\theta(x_i|x_{i-1},\ldots, x_{i-j+1})$. In practice, we consider the average perplexity on sequences of different lengths from the same dataset a suitable proxy for this.

Given $p_\theta$ and $L_t$, the problem that we want to solve is to develop
resource efficient methods that allow $p_\theta$ to extrapolate to input contexts of length $L_e$ that are longer than $L_t$. We refer to $L_e$ as the \emph{extended input
context length}.

%========================================================================
\subsection{Extending APE via interpolation}
\label{ssec:interpolate}
%========================================================================
%{{{1

APEs learn an embedding vector for each position up to a pre-specified maximum
position. The fixed nature of the embedding matrix does not allow for inputs longer
than the maximum pre-specified length. A necessary first step when training on longer
sequences is to increase the size of the embedding matrix.

We use linear interpolation to extend the embedding matrix from the training input
context length $L_t$ to the new input context length
$L_e$~\cite{Dehghani2023ScalingVT}. Let $E$ and $E'$ be the old and new embedding
matrices, respectively and assume that $\beta = L_e/L_t$ is integral. Then the
embedding for position $i$ ($0\le i < L_e)$ is given by:
\[
e'_i = \frac{\beta-i\%\beta}{\beta}e_{\lfloor i/\beta \rfloor} +
\frac{i\%\beta}{\beta}e_{\lfloor i/\beta\rfloor+1},
\]
where `\%' is the modulo operation. This process retains the original embeddings but results in $\beta (L_t - 1) + 1$ embeddings. In practice, we set the remaining $\beta - 1$ embeddings to $e_{L_t}$.

%========================================================================
\subsection{Efficient input context extension}
\label{ssec:sparse}
%========================================================================
%{{{1

Pairwise attention is the mechanism by which transformer models incorporate
information from other tokens. Positional embeddings are how attention takes into
account the absolute or relative positions of the token-pairs. To fully take
advantage of an increased input context, a model needs to learn the embeddings of the
newly created absolute positions or the relative embeddings associated with the
longer pairwise distances created with the increased input context. Thus, the model
needs to be further pre-trained with input sequences that also include the new
positions---in the case of absolute positional embeddings, or the longer pairwise
distances---in the case of relative positional embeddings.

The key insight behind our efficient approaches is that we can meet the above
requirements without directly training on long input sequences. Instead, we create
short input sequences by sampling segments from the long sequences, \emph{keep the
original positional information}, concatenate them, and use them to further pre-train
the language model. 
% These short input sequences are essentially gaped versions of the original long sequence. 
Since this approach retains the original positional
information, the models see the new positions/distances and learn how to use them.
Though the length of the short sequence is a hyper-parameter of our approach, in all
of our experiments we keep it the same as that of the original input context length;
i.e., $L_t$.

We develop two different subsequence sampling approaches that we refer to as
\chunk and \prefix which are defined as follows:
\begin{itemize}
    \item \textbf{\chunk-$\alpha$:} This approach creates a short sequence by sampling
    a small number of equal-length contiguous subsequences from the long sequence.
    Specifically, given $0<\alpha<1$ and an $L_e$-long input sequence $\mathbf{x}$,
    this approach samples $1/\alpha$ contiguous non-overlapping subsequences of
    length $\alpha L_t$ from $\mathbf{x}$. The reason that we keep the sampled
    segments contiguous is to preserve the local context information, which is
    important for next-token prediction~\cite{Xiong2021SimpleLA} and we do not want our model to
    `unlearn' it.
    \item \textbf{\prefix-$\alpha$}: This approach creates a short sequence by
    randomly sampling a set of tokens that forms a prefix and a contiguous segment to
    form its associated suffix. Specifically, given $0<\alpha<1$ and an input
    sequence $\mathbf{x}$ of length $L_e$, it randomly selects an index $i$ with
    $(1-\alpha)L_t < i < L_e-\alpha L_t$. It creates the suffix by taking the $\alpha
    L_t$ contiguous tokens starting at position $i$ and creates the prefix by
    randomly sampling $(1-\alpha)L_t$ tokens form the positions preceding $i$.
    In this method we only compute the loss over the continuous suffix in order to
    preserve the model's ability to incorporate local context.
\end{itemize}
A visualization of the different sampling methods can be found in Figure \ref{fig:fcm}.

%% Our proposed training method is a simple modification to the standard casual language
%% model training process. We sample a subset of $L_t$ tokens from the original $L_e$
%% long sequence while \textit{maintaining their original positions}.  This allows the
%% model to maintain the relative position between sampled tokens during training. For
%% absolute positional embeddings this enables the model to learn interactions between
%% positions greater then $L_t$ apart, for relative positional embeddings this decreases
%% the effect the tokens have on the following tokens.

While these methods can introduce discontinuities in the causal language modeling
objective we argue that maintaining their original positional embedding on top of the
fact they happen infrequently limits the harm they may cause. In practice we use $\alpha$'s small enough that discontinuities occurs approximately $2\%$ of the time in \chunk and never in \prefix.

%}}}1

%% Given two adjacent positional embeddings, $p_i, \; p_{i+1}$, and a duplication
%% factor, $f = {L_e}/{L_t}$, we extend to $ f - 1 $ new embeddings between $p_i$ and
%% $p_{i+1}$:
%%
%% \begin{multline}
%%     \frac{f-1}{f} p_i + \frac{f - (f - 1)}{f} p_{i+1},  \\
%%     \frac{f-1}{f} p_i + \frac{f - (f - 1)}{f} p_{i+1}, \cdots \\
%%     \frac{f-(f-1)}{f} p_i + \frac{f-1}{f} p_{i+1}
%% \end{multline}
%%
%% We perform this extension for each set of adjacent positions in the embedding matrix.
%% Including the original embeddings, this process results in an embedding matrix with
%% $k * L_t - k$ entries. We append the final embedding, $p_{L_t}$, $k$ times to the
%% embedding matrix to produce the final extended matrix with $kL_t$ positions. Once
%% the positional embedding matrix has been extended the model can process inputs of
%% length $kL_t$.

%}}}1

%% file: sections/5_experiment.tex
\section{Experimental setup}
\label{sec:exp}

\subsection{Dataset}
\label{ssec:dataset}

Since we are comparing the performance of various methods on long sequences we chose to use the scientific papers section of the arXiv dataset released by~\citet{Cohan2018ADA}. Scientific papers are a common choice for reporting results on long sequence modeling performance~\cite{Beltagy2020LongformerTL}.
This dataset consists of 215K scientific papers, split into 205K train and 7K test, with a total token count of approximately 1.6 billion and an average document length of 4,938 tokens. We do not pack our batches~\cite{Kosec2021PackingT2}, meaning each sequence contains only text from a single document at a time. 
If documents are longer than $L_e$ we split them into non-overlapping sequences with length $L_e$ and discard the remainder; if documents are shorter than $L_e$ we discard them as well. We feel that ensuring each input only corresponds to one source text is an important factor when reporting performance on long sequences. 

\subsection{Models}
\label{ssec:models}

To evaluate our methods we fine-tune three different classes of pretrained language models, one for each of the popular positional embedding methods: absolute, RoPE, and ALiBi. We use models with approximately 1.5 billion parameters; for absolute positional embeddings we use GPT-2~\cite{Radford2019LanguageMA}, for rotary embeddings we use Pythia ~\cite{Biderman2023PythiaAS}, and for ALiBi we use Bloom~\cite{Scao2022BLOOMA1}. In addition to these three models we use a smaller GPT-2 and Pythia checkpoint (approx. 10\% the size), which we will refer to as GPT-2 Small and Pythia Small, and together as our development models. Due to a lack of small models trained with ALiBi we do not have a development model for ALiBi. Key information about these models can be found in Table~\ref{tab:stat}.
Note that besides the positional encoding schemes, these models also differ in other ways including training data and model parameters. As a result, a direct comparison of these models will be confounded by these additional factors. For this reason our evaluation only focuses on measuring how the different continuous pre-training approaches help in improving each model's extrapolation capabilities against themselves and we never compare across models.

%While these five models vary in training data and total parameter count their results are never compared against each other. Instead we are only interested in measuring their improvement on extrapolation. 

\begin{table}[t]
\centering
  \caption{Key model characteristics.}
  \begin{tabular}{crrr}
    \toprule
    & \# of params & PE & $L_t$ \\
    \midrule
    GPT-2 Small & 170M & APE & 1024 \\
    Pythia Small & 140M & RoPE & 2048 \\
    \midrule
    GPT-2 & 1.64B & APE  & 1024 \\
    % Pythia & 1.4B  & RoPE & 2048 \\
    Pythia  & 1.4B & RoPE & 2048 \\ 
    Bloom & 1.45B  & ALiBi & 2048 \\
  \bottomrule
\end{tabular}
\label{tab:stat}
\end{table}

\subsection{Domain adaptation}
\label{ssec:da}

% The arXiv dataset is relatively out-of-domain for these models, with the models having high perplexity on the test set. 
The perplexity on arXiv for these models is relatively high as arXiv is considered out of domain. 
In order to differentiate between gains attributed to adapting to the domain versus improving extrapolation performance we perform one full epoch of continual pre-training with a sequence length of $L_t$ for each model. 

We refer to the checkpoints after domain-adaptation as "out-of-the-box" models.
% "Out of the Box" (OOTB) refers to models before any continual pre-training on new sequence lengths.
All experiments start from the OOTB models unless otherwise mentioned. The perplexity of the models after domain adaptation can be found in Table~\ref{tab:initial}.

\begin{table}[t]
 \centering
  \caption{Perplexity on sequences of the model's original input length, $L_t$, after domain adaptation.}
  \begin{tabular}{cr}
    \toprule
    & ppl. \\
    \midrule
    GPT-2 Small & 9.311 \\
    Pythia Small & 8.609 \\
    \midrule
    GPT-2 & 6.675  \\
    Pythia & 6.677  \\
    % Pythia & 5.730 \\ 
    Bloom & 7.217   \\
    % \midrule 
    % Pythia XL & 5.730 \\ 
  \bottomrule
\end{tabular}
\label{tab:initial}
\end{table}

\subsection{Segmented pre-training}
\label{ssec:segmented_pt}
% All other training details remain the same as the standard pretraining objective. 
For training we use the causal language modeling objective with a cross entropy loss. All experiments on the same model are done in a compute equivalent manner unless stated otherwise. To ensure compute equivalence when training our models we fix the number of tokens as well as the input length, $L_t$, of the model. 

Due to segmentation, one epoch of training on different sequence lengths results in a different number of tokens actually processed. For example, training with sequences of length $2L_{t}$ results in half the total number of tokens. To ensure an equal number of tokens across experiments we set the total number of epochs for each experiment to be:
\begin{equation}
    \mbox{\# epochs} = \frac{L_{e}}{L_{t}}.
\end{equation}

\subsection{Performance assessment}
\label{ssec:eval}

To evaluate the performance of our models on different sequence lengths we report the mean perplexity on sequence of length $L_e$ from our test set. Perplexity measures the exponentiated average negative log likelihood over a sequence of tokens and is a common evaluation metric for language models. We define the perplexity of a sequence of tokens \textbf{x} of length $L_t$ as:
\begin{equation}
    ppl(x) = \exp ( - \frac{1}{L_t} \sum_i \log p_\theta (x_i | x_{<i} )).
\end{equation}
Note that unlike previous work, we do not perform sliding window evaluation~\cite{Baevski2018AdaptiveIR}.

%% file: sections/6_results.tex
\section{Results}
\label{sec:results}
We conduct our experiments and present results in such a way to answer the following questions:
\begin{itemize}
    \item How well do absolute positional embeddings extrapolate with interpolation of the embeddings matrix?
    \item Which of our proposed subsequence sampling methods performs the best and with what parameters? 
    \item How does our approach compare with continual pre-training on sequences of the original length?  
\end{itemize}
\subsection{Out-of-the-box extrapolation}
\label{ssec:oob}

We begin by examining each model's ability to extrapolate to sequences longer then they were trained on without any further pre-training. We report the perplexity on the test set with sequence lengths starting from $L_{t}$ up to $5L_{t}$, depending on the memory constraints of each. Previous length extrapolation work did not include absolute positional embeddings due to their fixed nature~\cite{Press2021TrainST}. To increase the input context size we interpolated the positional embedding matrix as described in Section \ref{ssec:interpolate}. Results are shown in Figure~\ref{fig:out_of_the_box} and the corresponding numbers can be found in Table~\ref{tab:extrapolate_fig} in Appendix~\ref{sec:appendix}. 

\begin{figure}[t]
  % \centering
  \includegraphics[width=\linewidth]{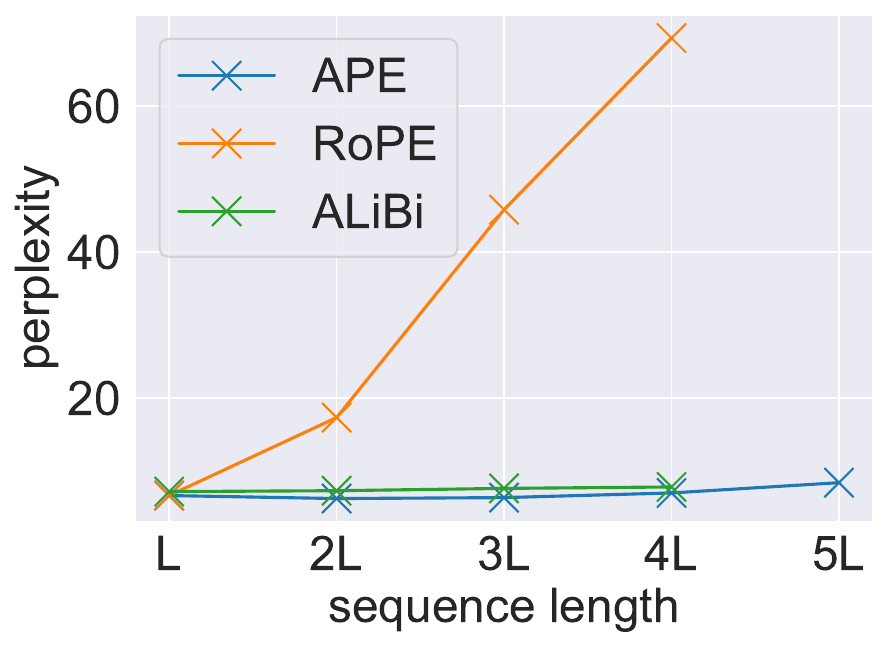}
  \caption{Perplexity of "out-of-the-box" extrapolation. With interpolation of the positional embeddings, absolute positional embeddings (APE) extrapolate as well as ALiBi. }
    \label{fig:out_of_the_box}
\end{figure}

RoPE fails to extrapolate to sequences longer than originally trained on while ALiBi generalizes well. These findings about RPEs agree with those previously observed in~\citet{Press2021TrainST}. Our results show that interpolation works well until at least $5L_t$. This suggests that with linear interpolation APEs generalize better than RoPE and are comparable to ALiBi.

%=============================================================================
\subsection{Comparison of segmented methods}
\label{ssec:small_res}
%=============================================================================

We compare the performance of the various methods discussed in Section~\ref{ssec:sparse} on our development models. We train models on two separate extension sizes, $L_e = 2 L_t$ and $L_e = 4 L_t$. For each we use \chunk with $\alpha=\{0.125, 0.25, 0.5\}$ and \prefix with $\alpha=\{0.25, 0.5 \}$. Furthermore, we train a models on sequences of $2L_t$ and $2L_t$ \textit{without} any segmentation. We refer to these models as \full, and they provide a point of comparison between our methods versus training on the full $L_e$ sequence. The complete set of results can be found in Table~\ref{fig:comparison}.

The different segment-based methods work well to extend the input context of these models. We observe a decrease in perplexity when evaluating on sequences longer then originally trained on. Overall, \chunk performs better than \prefix on both models, \prefix fails to improve extrapolation when extending RoPE to sequences $4\times$ in length. While the \full approach has the lowest perplexity in most cases the relative loss in performance for \chunk is low. One notable case is extending RoPE to $4L_t$, there we observe \chunk outperforming \full. Given that \chunk requires half the sequence length of \full it remains a competitive option due to its memory efficiency. 

Comparing the performance of different chunk lengths, controlled by the parameter $\alpha$, both models display similar trends. For \chunk, there appears to be sweet-spot between the number of segments and each segment's individual length (see Table~\ref{fig:comparison}). An $\alpha$ of 0.125 translates to chunks of 128 tokens for APE and 256 for RoPE. In most cases this $\alpha$ performed the worst amongst \chunk, as the segments may be too short or lead to too many discontinuities in the sequence. For \prefix, there is less of a concrete pattern. This could be due to the higher level of randomness in the prefix as tokens were sampled randomly. Between \chunk and \prefix, \chunk computes loss over twice as many tokens, this could be a contributing factor to the gap in performance between the two.

% Comparatively, 
Between RoPE and APE, RoPE benefits the most from segmented pre-training. 
% As highlighted in Section \ref{ssec:oob}, RoPE fails to extrapolate. 
After training on segmented sequences the perplexity on extensions of $2L_t$ and $4L_t$ decreases by a factor of $4\times$ and $24\times$ respectively. While our method still improves over the "out-of-the-box" performance of APEs, interpolation is a competitive approach for length extension.  

\begin{table}[t]
  \centering
    % \begin{threeparttable}

  \caption{Perplexity of different input context length extension methods on the development sets.}
  
  \begin{tabular}{llrr}
    \toprule
    & method & $2L_t$ & $4L_t$ \\
    \midrule
     \multirow{7}{*}{APE} & \OOTB & 9.322 & 13.275 \\ 
     & \full & 8.287 & 7.819 \\
     \cline{2-4}
     & \chunk-0.125 & 8.521 & 8.307   \\
     & \chunk-0.25  & 8.471 & \textbf{7.989}   \\
     & \chunk-0.5   & \textbf{8.420} & 8.259   \\
     & \prefix-0.25 & 8.757 & 8.826   \\
     & \prefix-0.5  & 8.672 & 9.304   \\
    \midrule
    \multirow{7}{*}{RoPE}  & \OOTB & 30.686 & 176.244 \\
     & \full & 7.403 & 7.353 \\
    \cline{2-4}
     & \chunk-0.125 & 7.476 & 7.239   \\
     & \chunk-0.25  & \textbf{7.447} &  \textbf{7.210}  \\
     & \chunk-0.5   & 7.461 & 7.461 \\
     & \prefix-0.25 & 9.543 &  25.539  \\
     & \prefix-0.5  & 10.119 &  33.375 \\
  \bottomrule
\end{tabular}
\label{fig:comparison}
    % \begin{tablenotes}
    %   \small
    %   \item Results unavailable for empty entries due to compute limitations. 
    % \end{tablenotes}
  % \end{threeparttable}
\end{table}

%=============================================================================
\subsection{Results on  larger models}
\label{ssec:big_res}
%=============================================================================

Based off the findings in Section~\ref{ssec:small_res} we use \chunk-$0.25$ for our experiments on GPT-2 1.5B, Pythia-1.4B, and, Bloom-1.1B. As before, we continually pre-train the models as detailed in Section~\ref{ssec:segmented_pt} and expand to $L_e = 2 L_t$ and $L_e = 4 L_t$. 

Overall, \chunk works for all three models on both expansion lengths. All models extrapolated better than their "out-of-the-box" performance. Again, RoPE was able to extrapolate to sequences it previously was not able to. Our method also demonstrated the ability to further increase the extrapolation ability of ALiBi. Results can be found in Table~\ref{tab:billion}.

\begin{table}[t]
  \centering
  \caption{Perplexity results for the 1.x billion parameter models.}
  \begin{tabular}{llrr}
    \toprule
    & method & $2 L_t$ & $4 L_t$ \\
    \midrule
    \multirow{3}{*}{APE} & \OOTB & 6.326 & 7.099   \\
         & {\tt DA} & \textbf{6.125}  & 7.050 \\

    & \chunk-0.25 & 6.314 & \textbf{6.425} \\ 
    \midrule
    \multirow{3}{*}{RoPE} & \OOTB & 16.428 & 52.644   \\
    & {\tt DA} & 16.285 & 50.652  \\
    & \chunk-0.25 & \textbf{5.448} & \textbf{5.278} \\
    \midrule 
    \multirow{3}{*}{ALiBi} & \OOTB & 7.295 & 7.773\\
         & {\tt DA} & 6.887  & 7.417  \\

    & \chunk-0.25 & \textbf{6.773} & \textbf{7.295} \\
  \bottomrule
\end{tabular}
\label{tab:billion}
\end{table}

%=============================================================================
\subsection{Comparison with further pre-training}
\label{ssec:further_pretraining}
%=============================================================================

Given that ALiBi and APE-based models already extrapolate well (see Figure~\ref{fig:out_of_the_box}), a natural question is whether the performance gains on longer sequences come from our segmented method or additional domain adaption. To ablate this, we perform another epoch of domain adaptation as described in Section~\ref{ssec:da}. This isolates the benefit of our method versus further domain adaptation as the total number of tokens seen by all models are the same. Results can be found in Table~\ref{tab:billion}. 

For models that extrapolate well (ALiBi and APE), further domain adaptation also improves the extrapolation ability however the gains are less than our segmented training. The exception here is when extending APE to lengths $2\times$, in this case domain adaption performs slightly better. This result indicates that the interpolation-based extension method we propose works well for APEs. Overall, this demonstrates that while some of the gains may be due to further domain adaptation our method is still beneficial for models that extrapolate well "out-of-the-box".

% \begin{table}[t]
%   \centering
%   \caption{Comparison of ALiBi's extrapolation with further domain adaptation (DA) versus our method.}
%   \begin{tabular}{llrr}
%     \toprule
%     & method & $2 L_t$ & $4 L_t$ \\
%      \midrule
%     \multirow{3}{*}{APE} & \OOTB & 6.326 & 7.099    \\
%      & {\tt DA} & 6.125  & 7.050 \\
%      & \chunk-0.25 & 6.314 & 6.425 \\

%     \midrule
%     \multirow{3}{*}{ALiBi} & \OOTB & 7.295 & 7.773   \\
%      & {\tt DA} & 6.887  & 7.417  \\
%      & \chunk-0.25 & 6.773 & 7.295 \\
%   \bottomrule
% \end{tabular}
% \label{tab:da_2}
% \end{table}

%=============================================================================
\subsection{Comparison with RandomPos}
\label{ssec:ablation}

The authors of \randpos~\cite{Ruoss2023RandomizedPE} proposed a similar method for simulating training on long sequences within a fixed input context window. Instead of subsampling sequences of length $L_e$, \randpos randomized the positional ids of sequences of length $L_t$ selecting positions ranging from $[0, L_e - 1]$ while maintaining the causal ordering. Similar to our approach, \randpos exposes the model to extrapolated pairwise relative distances but the key difference is content used. Whereas \randpos only presents local context to the model, \chunk exposes the model to distant content and encourages the model to learn to leverage distant contexts.

To verify the exposure to distant content is an important step in improving extrapolation we implement a version of \randpos and extend our models to $2\times$ and $4\times$ the original input sizes. We keep all settings and models the same as Section~\ref{ssec:small_res} with the exception of including the ALiBi model. In all cases, \chunk outperforms \randpos indicating the inclusion of distant context valuable to length extrapolation. Results can be found in Table~\ref{tab:ablation}. 
%=============================================================================

\begin{table}[t]
  \centering
  \caption{Comparision with \randpos. Numbers reported are perplexity.}
  \begin{tabular}{llrr}
    \toprule
    & method & $2 L_t$ & $4 L_t$ \\
    \midrule
    \multirow{3}{*}{APE} & \OOTB & 9.322 & 13.275   \\
         & \randpos & 9.018  & 11.534 \\
    & \chunk-0.25 & \textbf{8.420} & \textbf{7.989} \\ 
    \midrule
    \multirow{3}{*}{RoPE} & \OOTB & 30.686 & 176.244   \\
    & {\randpos} &  8.021 & 11.692 \\
    & \chunk-0.25 & \textbf{7.447} & \textbf{7.210} \\
    \midrule
    \multirow{3}{*}{ALiBi} & \OOTB & 7.295 & 7.773\\
     & {\tt DA} & 6.816  & 7.352  \\
    & \chunk-0.25 & \textbf{6.773} & \textbf{7.295} \\

  \bottomrule
\end{tabular}
\label{tab:ablation}
\end{table}

%% file: sections/7_discussion.tex
%========================================================================
\section{Analysis}
\label{sec:disc}
%========================================================================

Our results demonstrate that segmented training is a viable approach to extend the input context size of language models. It is not immediately intuitive why, especially given that the relative positional embeddings methods are not learned. 

For absolute positional embeddings the reasoning is fairly straightforward. First, in Section~\ref{ssec:oob} we demonstrated interpolating the embedding matrix led to reasonable extrapolation without any training. Before any training occurs the model already has some extrapolation ability. The segmented sequences allow for positions further away than the input size normally allows to interact and learn how to incorporate information. 

% \cite{Han2023LMInfiniteZE}
In the case of relative positional embedding methods these results are less intuitive. Both RPE methods penalize the attention scores of positions as a function of their relative distance, meaning that initially there is not much attention across chunk boundaries. We hypothesize that through training on segmented sequences the model learns to attend to longer-range interactions. There is a lack of nearby positions for the model to attend to so it learns to incorporate information from further away. In doing so it adjusts the weights to penalize further positions less. This counteract-acts the RPE's inductive bias towards nearby positions. 

\begin{figure}[t]
    \centering
    \includegraphics[width=\linewidth]{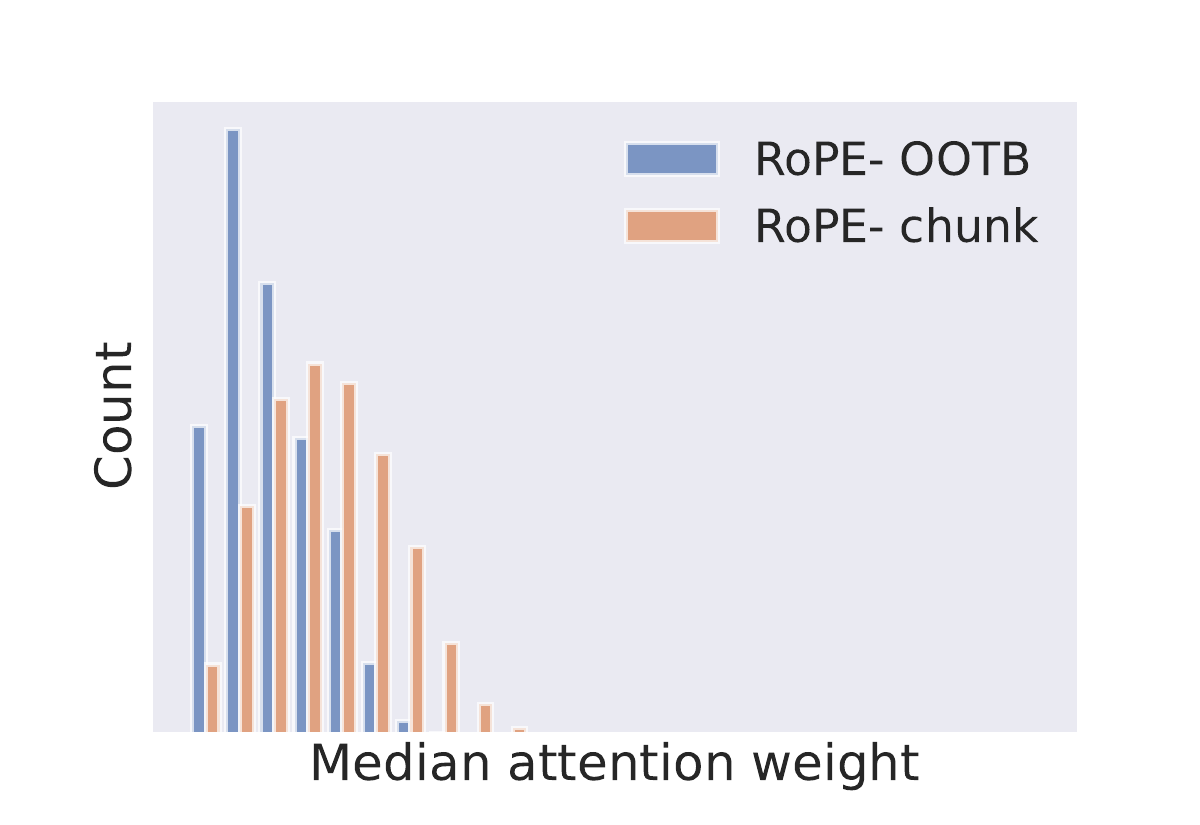}
    \caption{Histogram of median attention weights for positions past the original input length before and after our segmented training on models with RoPE. After adaptation, the distribution of attention weights becomes more uniform.}
    \label{fig:rope_after}
\end{figure}

To attempt to visualize this we plot the distribution of median attention weights for positions past $L_t$. In both cases, the medians are well below the mean suggesting that a few positions account for the majority of the attention weight. After segmented training, we observe the average median increases as well as become more evenly distributed. This suggests that more positions are being attended to as well as the model attending to more or less positions depending on the context. The plot can be found in Figure~\ref{fig:rope_after}. This hypothesis is also supported by a recent work that analyzes the failure of RoPE to generalize to long sequences~\cite{Xiong2023EffectiveLS}. The observed that simply reducing the decaying effect of RoPE distant tokens lead to strong extrapolation performance. 
% This explanation is partially supported by previous work. The authors of~\citet{Kazemnejad2023TheIO} reported that Transformer-based decoder language models generalized to variable sequence lengths best when trained with no positional embeddings. They demonstrate that the self-attention layers can learn both absolute and relative positional embeddings when trained with stochastic gradient descent. It has also been demonstrated that language models can be trained entirely without positional embeddings~\cite{Haviv2022TransformerLM} further suggesting a more uniform positional bias is beneficial. 

% Similarly, the forgetful causal masking~\cite{Liu2022FCMFC} method, which randomly masked out a set of previous tokens, demonstrated an increase in few-shot and fine-tuning performance. The authors argued their method of masking out a subset of tokens discouraged the model from over-attending to certain tokens in the input. Our method of segmenting long sequences could be realizing some of the same benefits. 

%% file: sections/8_conclusion.tex
\section{Conclusion}
\label{sec:conclusion}

In this work we proposed a simple and memory efficient approach to extend the effective input context size of models through training on sequences created by sampling segments from long documents. We demonstrated our method is robust to the choice of positional embeddings and allows models to be trained on sequences at least $4 \times$ their original input length. Furthermore, our results on extending absolute positional embeddings through interpolation demonstrated they can extrapolate better than RoPE and provide a method to extend the context of models trained with APEs at no additional cost. 

%% file: sections/9_limitations.tex
\section{Limitations}

In this work we explore various computationally efficient methods for pre-training on long sequences. 
Due to the compute limitations we only verify our method's performance on models up to 1.4 billion parameters. 
Current state of the art models are orders of magnitudes larger. While our results indicate the success of our method there is always the chance that results do not transfer to different model sizes. We believe these methods will hold as model size increases since the extrapolation problem is fundamentally an artifact of the positional embeddings and not model size. Additionally, the models we used were originally only trained with a maximum sequence length up to 2048 tokens and only extended to a maximum 8192 tokens. Even though this is a $4 \times$ extension, this is much lower then the input size of some production models.

Inline with previous work on encoding positional information~\cite{Press2021TrainST, Su2021RoFormerET}, we use perplexity as our method for evaluating a model's extrapolation performance. Some recent work has shown that this may not always be a strong signal for downstream performance~\cite{shaham-etal-2022-scrolls}. A more thorough evaluation on downstream benchmarks would be insightful, unfortunately the majority of our models were too weak to produce competitive performance on zero-shot or few-shot long sequence tasks. 

% There remain many active directions with regard to our segmented sequences approach. Some promising directions include curriculum based training, further expansions, and mixing different sequence lengths during training. Our method can be applied iteratively in a curriculum learning setting, increasing the model's input length at iteration. This could potentially allow for better performance on the final context length. Incorporating different sequences lengths during training, whether it be sequences of length $L_t$ and $L_e$ or a mix of different extended lengths could lead to further extrapolation gains. Incorporating different sequence lengths could ensure the model size does not over-fit to any particular sequence length and including non-segmented sequences could ensure the model does not lose the ability to capture local context. Finally, due to memory constraints we were unable to evaluate the performance on sequences greater then $4L_t$ on the 1.5 billion parameter models. The larger the extension the more efficient our method becomes. Identifying the upper bound of possible extension length of our method is important to understanding its practicality. 

%% file: sections/10_ethics.tex
\section{Ethics statement}

When working with language models and large, web-crawled datasets it is important to remain cognizant of some of the potential ethical concerns. We trained on scientific papers which are voluntarily posted by users. 
% The ability to extend input contexts and allow for longer generation could increase the issue of model generation

%% file: sections/11_appendix.tex
\section{Full Results}
\label{sec:appendix}

\begin{table}[t]
\centering
  \caption{Perplexity of "out of the box" extrapolation for models with APE, RoPE, and ALiBi positional embeddings.}
  \begin{tabular}{crrrrr}
    \toprule
    (ppl.) & $1\times$ & $2\times$ & $3\times$ & $4\times$ & $5\times$ \\
    \midrule
    APE  & 6.675	&6.326	&6.394	&7.099	&8.438 \\ 
    RoPE  & 6.677 &	17.348	& 45.797	& 69.288 &	- \\
    ALiBi & 7.217 & 7.295	& 7.653	& 7.773	& - \\
  \bottomrule
\end{tabular}
\label{tab:extrapolate_fig}
\end{table}